\documentclass[10pt,twocolumn,letterpaper]{article}

\usepackage[pagenumbers]{cvpr} 

\usepackage{multirow}
\usepackage{lineno}
\usepackage{fontawesome5}
\usepackage{twemojis}
\usepackage{xcolor}

\usepackage{comment}

\definecolor{cvprblue}{rgb}{0.21,0.49,0.74}
\usepackage[pagebackref,breaklinks,colorlinks,allcolors=cvprblue]{hyperref}

\newcommand{\Sheep}[1][]{\includegraphics[width=10pt,trim={6cm 7cm 5cm 6cm},clip]{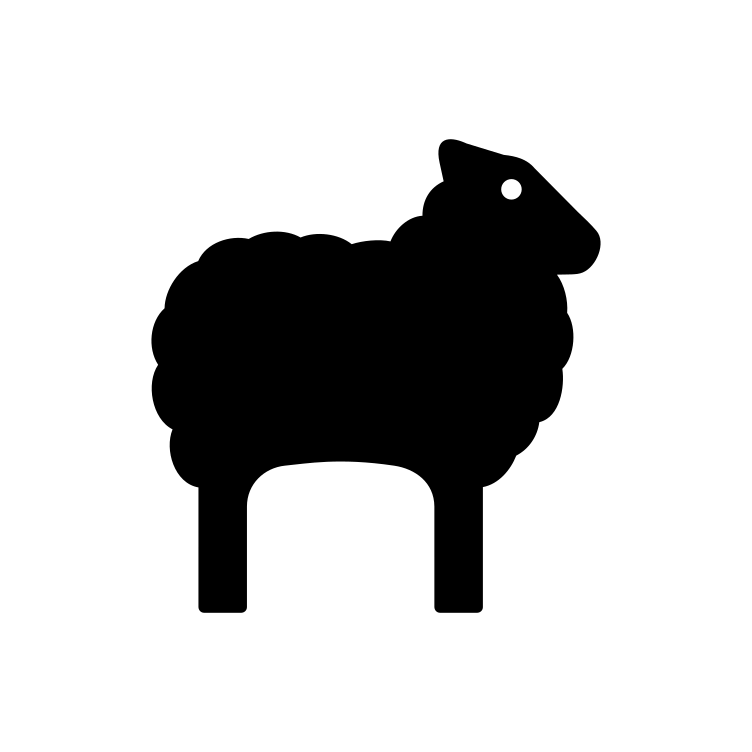}}
\newcommand{\Backpack}[1][]{\includegraphics[height=10pt]{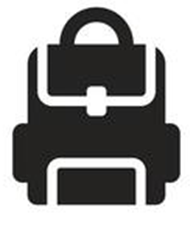}}
\newcommand{\Bench}[1][]{\includegraphics[height=10pt]{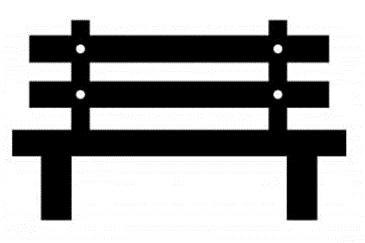}}
\newcommand{\Hairdryer}[1][]{\includegraphics[height=10pt]{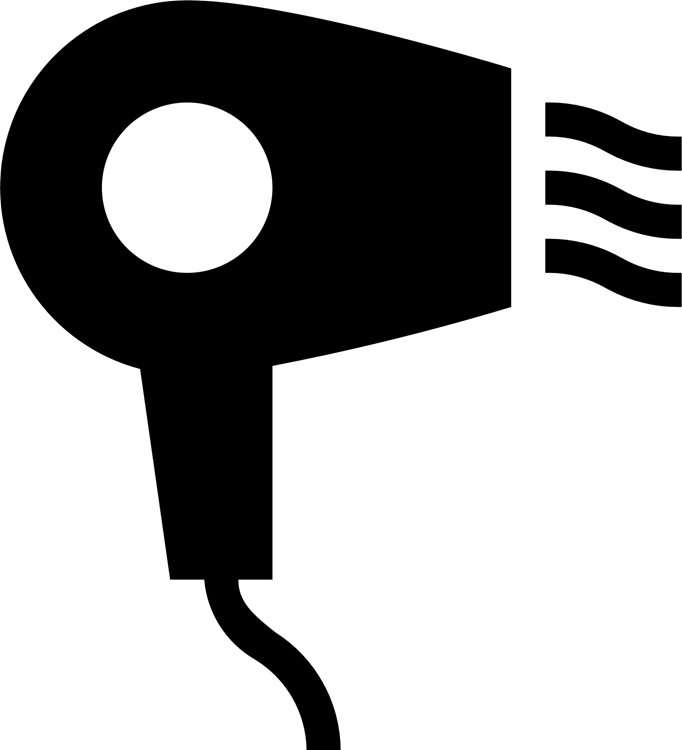}}
\newcommand{\Microwave}[1][]{\includegraphics[height=10pt]{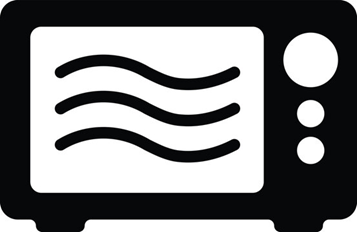}}
\newcommand{\Mouse}[1][]{\includegraphics[height=10pt]{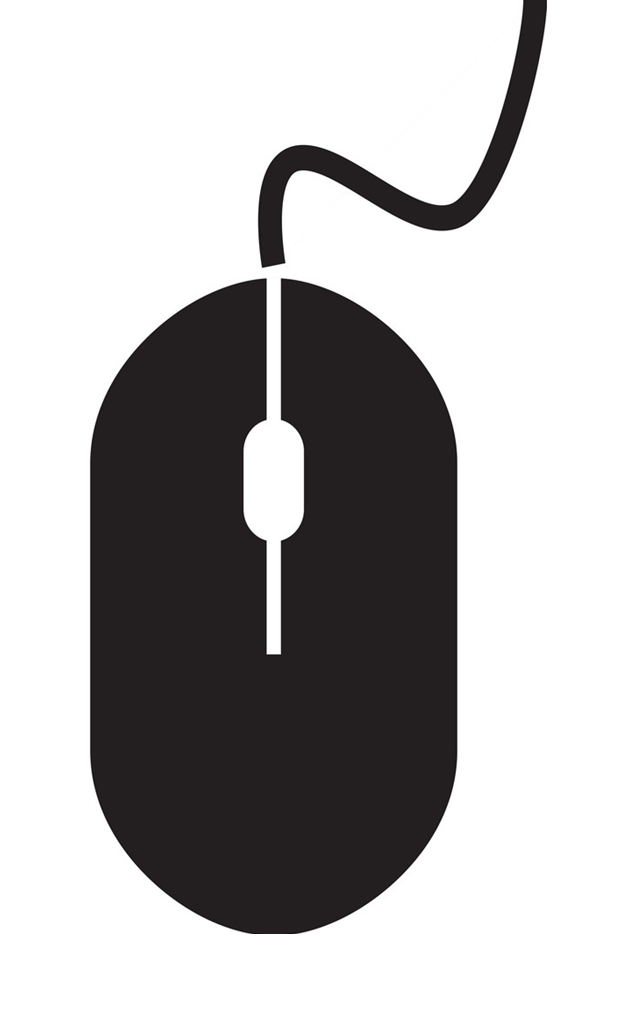}}
\newcommand{\Remote}[1][]{\includegraphics[height=10pt]{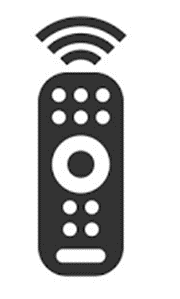}}
\newcommand{\Toaster}[1][]{\includegraphics[height=10pt]{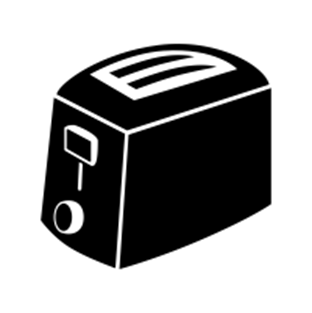}}
\newcommand{\Toilet}[1][]{\includegraphics[height=10pt]{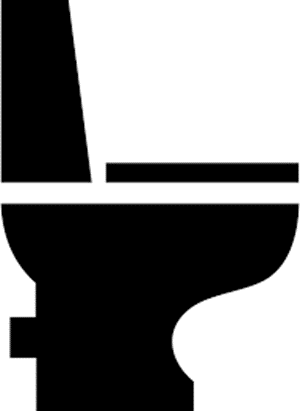}}

\def\paperID{7572} 
\def\confName{CVPR}
\def\confYear{2025}
\newcommand{\OURS}{Common3D\xspace}

\title{\OURS: Self-Supervised Learning of 3D Morphable Models \\for Common Objects in Neural Feature Space}

\author{Leonhard Sommer 
\textsuperscript{1}
\and
Olaf Dünkel \textsuperscript{2}
\and
Christian Theobalt \textsuperscript{2}
\and
Adam Kortylewski \textsuperscript{1,2}
\and
\\
\textsuperscript{1}University of Freiburg \quad \textsuperscript{2}Max Planck Institute for Informatics
}

\begin{document}
\twocolumn[{%
\renewcommand\twocolumn[1][]{#1}%
\maketitle
\begin{center}
    \centering
    \captionsetup{type=figure}
    \includegraphics[width=\textwidth]{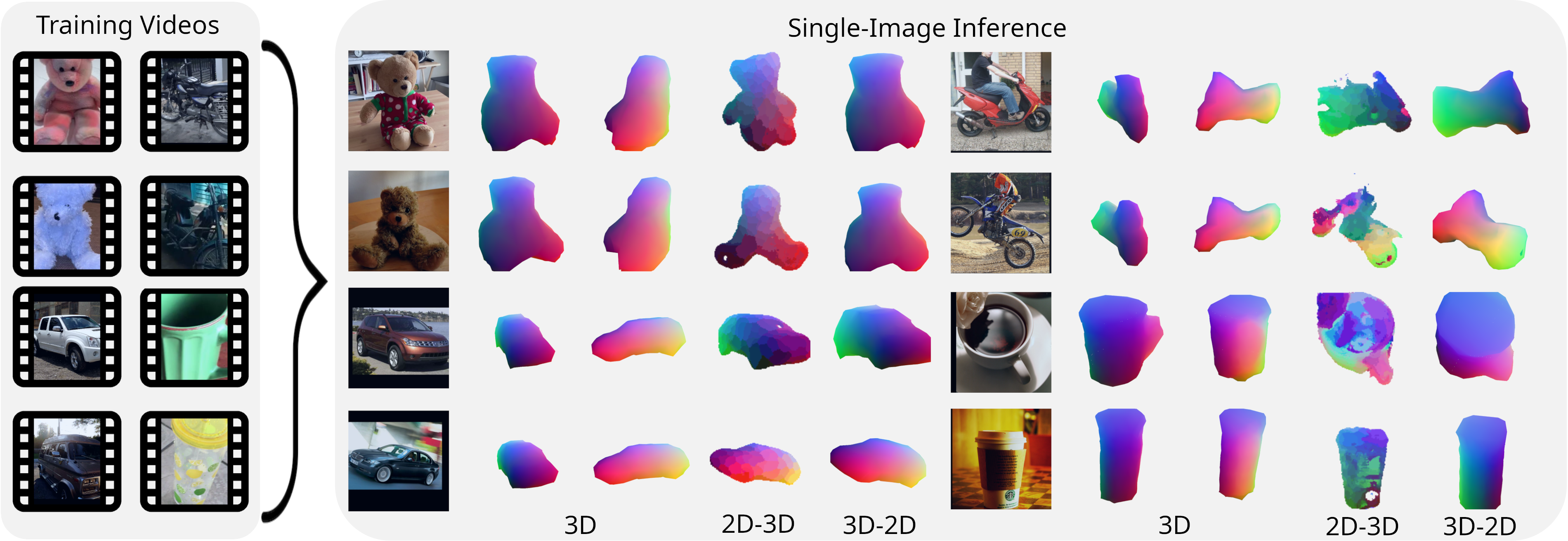} 
    \captionof{figure}{\textbf{Common3D} learns category-specific 3D morphable models from few casually captured videos completely self-supervised, and can estimate the 3D object shape (visualized from two viewpoints), 2D-3D correspondences, and the 3D object pose via inverse rendering.}
    \label{fig:teaser} 
\end{center}%
}]
\begin{abstract}
3D morphable models (3DMMs) are a powerful tool to represent the possible shapes and appearances of an object category.
Given a single test image, 3DMMs can be used to solve various tasks, such as predicting the 3D shape, pose, semantic correspondence, and instance segmentation of an object.
Unfortunately, 3DMMs are only available for very few object categories that are of particular interest, like faces or human bodies, as they require a demanding 3D data acquisition and category-specific training process.
In contrast, we introduce a new method, \OURS, that learns 3DMMs of common objects in a fully self-supervised manner from a collection of object-centric videos.
For this purpose, our model represents objects as a learned 3D template mesh and a deformation field that is parameterized as an image-conditioned neural network.
Different from prior works, \OURS represents the object appearance with neural features instead of RGB colors, which enables the learning of more generalizable representations through an abstraction from pixel intensities.
Importantly, we train the appearance features using a contrastive objective by exploiting the correspondences defined through the deformable template mesh. 
This leads to higher quality correspondence features compared to related works and a significantly improved model performance at estimating 3D object pose and semantic correspondence.
%
\OURS is the first completely self-supervised method that can solve various vision tasks in a zero-shot manner. We release all code at \href{https://github.com/GenIntel/common3d}{github.com/GenIntel/common3d}.
\end{abstract}    
\newcommand{\MM}{Common3D}
\newcommand{\MMs}{Common3Ds}

\newcommand{\featf}{\phi_{feat}}
\newcommand{\affinef}{\phi_{a}}
\newcommand{\sdff}{\phi_{sdf}}

\newcommand{\affinefscale}{\alpha}
\newcommand{\affineftransl}{\delta}

\newcommand{\vertexsymbol}{\mathrm{v}}

\newcommand{\vertex}{\vertexsymbol}
\newcommand{\vi}{\vertexsymbol_i}
\newcommand{\vj}{\vertexsymbol_j}
\newcommand{\vk}{\vertexsymbol_k}

\newcommand{\feati}{\feat_i}
\newcommand{\featj}{\feat_j}
\newcommand{\featr}{\feat_r}
\newcommand{\featk}{\feat_k}
\newcommand{\featmap}{\mathrm{f}}
\newcommand{\featmapi}{\mathrm{f}}
\newcommand{\featmapr}{\bar{\mathrm{f}}}
\newcommand{\featb}{\beta}

\newcommand{\instlatent}{\mathrm{l}}
\newcommand{\instlatentspace}{\mathrm{L}}

\newcommand{\featencf}{\Psi_f}
\newcommand{\featencr}{\Psi_r}
\newcommand{\featencb}{\Psi_b}
\newcommand{\featencl}{\psi_\instlatent}

\newcommand{\edges}{\mathrm{E}}

\newcommand{\prob}{\mathrm{p}}
\newcommand{\probpred}{\bar{\prob}}
\newcommand{\probrend}{\hat{\prob}}
\newcommand{\pose}{\pi}

\newcommand{\pointsinst}{\mathcal{P}_{inst}}
\newcommand{\pointssdf}{\mathcal{P}_{sdf}}

\newcommand{\mask}{\mathrm{m}}
\newcommand{\maskpred}{\bar{\mask}}
\newcommand{\maskdt}{\mask_{\mathrm{dt}}}

\newcommand{\height}{\mathrm{H}}
\newcommand{\width}{\mathrm{W}}

\newcommand{\lossmask}{\mathcal{L}_\mask}
\newcommand{\lossmaskdt}{\mathcal{L}_{\mask\mathrm{dt}}}
\newcommand{\losseik}{\mathcal{L}_{sdf}}
\newcommand{\lossrec}{\mathcal{L}_{CD}}
\newcommand{\lossregdef}{\mathcal{L}_{def}}
\newcommand{\lossregdefsm}{\mathcal{L}_{def-sm}}
\newcommand{\lossapp}{\mathcal{L}_{app}}

\newcommand{\glossmask}{\lambda_\mask}
\newcommand{\glossmaskdt}{\lambda_{\mask\mathrm{dt}}}
\newcommand{\glosseik}{\lambda_{sdf}}
\newcommand{\glossrec}{\lambda_{CD}}
\newcommand{\glossregdef}{\lambda_{def}}
\newcommand{\glossregdefsm}{\lambda_{def-sm}}
\newcommand{\glossapp}{\lambda_{app}}

\newcommand{\vv}{\vertexsymbol}
\newcommand{\vcyclce}{(\vv^{\mathrm{nn,f}})_{}^{\mathrm{nn,f}}}
\newcommand{\vnnfeat}{\psi{}} 
\newcommand{\vnneuc}{\chi{}}  
\newcommand{\vinnfeat}{\vertexsymbol_i^{\mathrm{nn,f}}}
\newcommand{\vinneuc}{\vertexsymbol_i^{\mathrm{nn,x}}}
\newcommand{\vjnnfeat}{\vertexsymbol_j^{\mathrm{nn,f}}}
\newcommand{\vjnneuc}{\vertexsymbol_j^{\mathrm{nn,x}}}

\newcommand{\vertexsrcsymbol}{\vertexsymbol_{\mathrm{src}}}
\newcommand{\vertexrefsymbol}{\vertexsymbol_{\mathrm{ref}}}
\newcommand{\featbgsymbol}{\mathrm{b}}

\newcommand{\vertexword}{vertex}
\newcommand{\Vertexword}{Vertex}

\newcommand{\imagesymbol}{\mathrm{I}}
\newcommand{\imageword}{Image}

\newcommand{\pixelsymbol}{\mathrm{p}}
\newcommand{\pixelword}{Pixel}

\newcommand{\encodersymbol}{\phi}
\newcommand{\encoderword}{Image Encoder}

\newcommand{\feats}{\mathrm{F}}
\newcommand{\feat}{\mathrm{f}}
\newcommand{\featdim}{\mathrm{D}}
\newcommand{\featword}{Feature}
\newcommand{\featsword}{Features}

\newcommand{\argmin}{\arg\!\min}
\newcommand{\verts}{\mathrm{V}}
\newcommand{\vertsdef}{\mathrm{V}_{def}}
\newcommand{\faces}{\mathcal{F}}
\newcommand{\mesh}{\mathrm{M}}
\newcommand{\meshdef}{\mathrm{M}_{def}}

\newcommand{\refverts}{\bar{\verts}}
\newcommand{\srcverts}{\verts}

\vspace{-.25cm}
\section{Introduction}
\label{sec:intro}
Estimating the 3D shape and pose of an object from a single image requires \textit{prior knowledge} about the possible shapes and appearances of an object category.
Learning such prior models is a long-standing problem in computer vision, with 3D morphable models (3DMMs) \cite{blanz1999morphable} emerging as a very successful approach to address this problem.
A popular application of 3DMMs is to perform single image 3D reconstruction via inverse rendering \cite{blanz1999morphable,aldrian2012inverse,tewari2017mofa}, also referred to as analysis-by-synthesis approach \cite{yuille2006vision}.
This enables solving under-constrained vision tasks, for example predicting an object’s shape behind an occluder \cite{egger2018occlusion} or estimating the 3D object shape and pose from a single image.
However, training 3DMMs requires sophisticated data acquisition setups \cite{anguelov2005scape,joo2018total,liu2013markerless}, involving laser scanners or multi-view studio captures, and a significant manual effort in cleaning and registration of the data \cite{blanz1999morphable,loper2015smpl,zuffi20173d}. 
This approach is difficult to scale to more common everyday objects. 
An alternative approach that emerged recently is to learn a 3D prior from 2D images only \cite{wu2020unsupervised,yang2022banmo,tewari2022disentangled3d,zhang2023seeing}.
However, learning such priors from in-the-wild images is a major challenge and typically requires significant category-specific design choices, with various solutions having been proposed for different object categories \cite{kanazawa2018learning,wu2023dove,wu2023magicpony,li2024learning,zhang2023seeing}.
This makes it difficult to scale to the long tail of common objects that can appear in natural images.

We introduce \textbf{\OURS}, a new way of integrating self-supervised learning, neural rendering, and 3D modeling to learn 3D object priors for common objects from a collection of object-centric videos.
Once trained, the model can estimate 2D-3D correspondences, it can be leveraged for inverse rendering to estimate the 3D pose, and instance segmentation of common objects from a single image, see \cref{fig:teaser}.
At the core of our method is a category-specific 3D morphable model (3DMM) that represents the possible variations of different object instances.
The 3DMM is represented using a learned prototypical 3D shape and a shape deformation field that is parameterized as an image-conditioned neural network.
Training our model requires canonical camera poses, which we estimate unsupervised by adopting the method of \cite{sommer2024unsupervised}.

One core challenge for learning 3DMMs of common objects is to be flexible enough to capture the highly variable shapes and topologies of various object categories, as well as the possible variations of instances within one category.
We address this issue by using a hybrid volumetric mesh representation based on DMTet \cite{shen2021deep}. 
Hence, the shape is defined volumetrically in canonical space, but a mesh is extracted on the fly for morphing and rendering the model. 
This allows to retain the flexibility of implicit neural representations for capturing highly variable shapes and topologies \cite{chen2019learning,mescheder2019occupancy,park2019deepsdf}, while also enabling the use of meshes for an efficient rendering and regularization during training. 

Another important challenge is to perform inverse rendering with 3D object priors, as it requires the optimization of a highly non-convex image-based reconstruction loss \cite{schonborn2017markov,tewari2017mofa,romaszko2017vision}.
We address this problem in \OURS by representing the object appearance as a semantic feature field instead of RGB colors.
Performing inverse rendering on the feature-level is easier because the features can directly encode 2D-3D correspondences.
While related work \cite{wu2023magicpony,li2024learning} directly uses DINO \cite{oquab2023dinov2} as appearance features, recent studies showed that DINO has severe limitations at estimating semantic correspondence accurately \cite{mariotti2024improving,zhang2024telling}.
Therefore, we train an additional adapter module on top of DINO by using the 3D object prior to further enhance the feature extractor at accurately estimating 2D-3D correspondences.
This leads to higher quality semantic correspondence features and a largely enhanced model performance.

To summarize, we make the following \textbf{contributions}: 
(1) We propose a new way of integrating self-supervised learning, neural rendering, and 3D modeling to enable the joint learning of 3D object priors for common objects from a collection of object-centric videos,
(2) an effective mechanism for learning correspondence features by training a 3D object prior jointly with an additional adapter module on top of DINO, using a local contrastive objective that exploits the correspondences estimated by the 3D model, and 
(3) \OURS is the first completely self-supervised method that can estimate the shape, pose, and semantic correspondence of common objects from a single image. 

\section{Related Work}
\label{sec:related}

Our model has two core components: An adapter module that enhances correspondence learning and a 3D morphable model that is learned without supervision.
Therefore, we first provide an overview of correspondence learning methods and subsequently discuss related work on learning 3D morphable models from images and from synthetic data.

\begin{figure*}
    \centering
    \includegraphics[width=1.\linewidth]{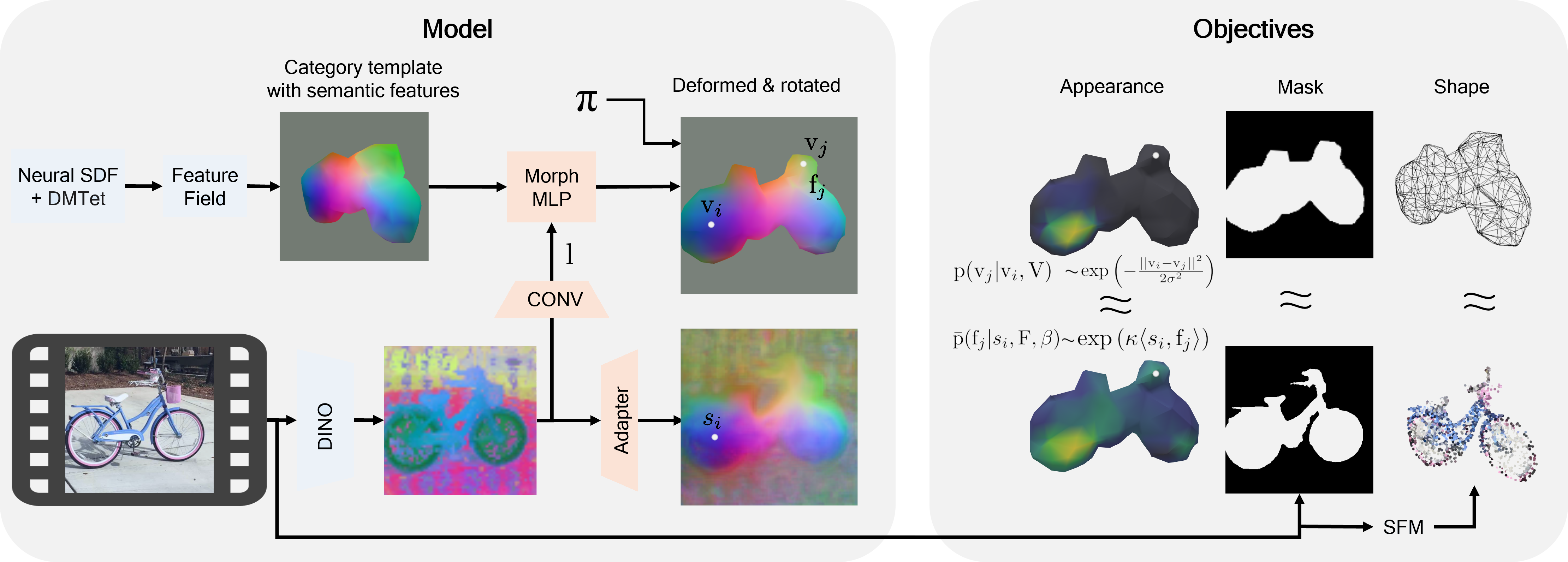}
    \caption{\textbf{Method Overview.} At the core of our method is a category-level template with semantic features that is acquired using a neural SDF with Differentiable Marching Tetrahedra (DMTet) where features are attached using a feature field. 
    The category-level template is morphed using an MLP that is conditioned on a latent code $\instlatent$ and rotated using the pose $\pose$. The pose $\pose$ is estimated unsupervised at training time and predicted at inference time.
    DINOv2 serves as a feature encoder and its output is processed in two branches: One for estimating the latent code $\instlatent$ with a block of convolutional layers; and another branch serves as an adapter to enhance the DINO features for better correspondence learning.
    As a result, the features from the adapter are better suited for finding correspondences. For example, the tire features of DINO have a similar encoding and are, hence, ambiguous, whereas it is much more distinct after the adapter.
    We also visualize the training objectives:
    The appearance objective compares two surface probabilities: 1) The geometric probability $\prob(\vj | \vi, \verts)$, as defined via the Euclidean distances of the mesh vertices; and 2) the appearance probability $\probpred(\featj | s_i, \feats, \featb)$, acquired by comparing the 2D image features $s_i$ with the surface features $\feats$.
    The remaining geometric objectives enforce the 3D shape and the projected mask to be consistent with the data acquired from a SFM pipeline.
    }
    \label{fig:method}
\end{figure*}

\textbf{Self-Supervised Correspondence Learning.}
Finding semantic correspondences across instances of an object category is a challenging computer vision task \cite{min2019spair}.
Self-supervised models like DINOv2 \cite{oquab2023dinov2} and Stable Diffusion \cite{rombach2022high} were shown to have remarkable capabilities at estimating zero-shot semantic correspondences \cite{zhang2024tale}. 
Despite these advances, unsupervised models are still not performing well on all categories.
This can be accounted to ambiguous features of similar parts \cite{mariotti2024improving} or the lower performance for viewpoint changes \cite{zhang2024telling}.
Previous works have addressed this by mapping the 2D features to a coarse spherical representation of the object \cite{mariotti2024improving} or by applying heuristic augmentation methods to account for the viewpoint dependency \cite{zhang2024telling}.
Recent related work \cite{shtedritski2024shic} showed that by leveraging 3D shape representations, the 2D-3D correspondence mapping can be improved significantly. 
However, their approach requires a detailed categorical mesh and is limited to a small selected set of animals.
Similarly, \cite{sommer2024unsupervised} uses multiple views from object-centric videos to account for the viewpoint dependency. Further, they obtain a simple rigid 3D template and learn 2D-3D correspondences to estimate the 3D pose of an object.
In contrast, our proposed \OURS learns an underlying deformable 3D shape representation self-supervised,
%
which enables us to generalize to various common objects.
Moreover, our approach of learning correspondence features by training a 3D object prior jointly with an additional adapter module on top of DINO enables high quality features to emerge.
\newpage
\textbf{Learning 3D Morphable Models from Images.}
There was remarkable progress made in the self-supervised learning of category-specific 3D morphable models from images. 
%
%
Originally, approaches used camera viewpoint, 2D keypoints, and masks to predict the 3D shape \cite{kar2015csm, tulsiani2016learning} and texture \cite{kanazawa2018cmr, wang2023dtf}. 
Several works achieved to mitigate the requirement of camera viewpoint and 2D keypoint annotations by exploiting symmetries \cite{goel2020ucmr, li2020umr, tulsiani2020imr, hu2021smr}, by adding multi-view viewpoint hypothesis \cite{goel2020ucmr}, by requiring consistent self-supervised co-part segmentation \cite{li2020umr} using \cite{hung2019scops}, by cycle-consistent surface maps \cite{tulsiani2020imr, kulkarni2020articulation, kokkinos2021ttp}, or by temporal consistency for video clips \cite{tewari2019fml,tewari2021learning,wu2023dove}. 
Few works even achieve to not require mask annotations, \cite{wu2020unsup3d, monnier2022unicorn,tewari2022disentangled3d}, by being constrained to front views of faces \cite{wu2020unsup3d}, or coarse-to-fine learning of texture and shape \cite{monnier2022unicorn}. 
However, most of these works focus on modeling the object in very fine detail, such that they can generate object instances down to the 
RGB pixel-level, making the inverse rendering process inherently difficult \cite{schonborn2017markov,tewari2017mofa,romaszko2017vision}.
In contrast, our focus lies on self-supervised learning of a 3D morphable model that is particularly well suited for solving vision tasks through inverse rendering. 
Therefore, we represent the object appearance using neural features instead of RGB colors, which facilitates the inverse rendering process
as the features learn to support the inverse rendering process by encoding 2D-3D correspondences.

Another line of work learns 3D object priors without viewpoint annotations by replicating DINO features \cite{yao2022lassie, yao2023hi, wu2023magicpony, li2024learning, jakab2024farm3d, sun2025ponymation}. 
In contrast to us, these works are restricted to animal categories. 
Further, they are not enhancing the image feature encoder for learning correspondences. 
The work of Zhang et al. is related to ours \cite{zhang2022self}, as it also enhances the image feature encoder, but not in an inverse rendering setting. 
Further, they rely on a detailed template prior shape and focus on categories with few deformations and rotational symmetries, such as a bottle. 


There exist also multiple works that focus on video-specific deformation model \cite{park2021nerfies, yang2021viser, yang2022banmo, das2024neural}. 
In contrast to us, they have to re-train for each new instance and cannot find correspondences across instances.

\textbf{3D Morphable Models from Synthetic Data.}
Various work use synthetic data \cite{chang2015shapenet} to learn 3D representations. 
With dense annotation, in terms of a complete surface and annotated canonical frames, several works learn impressive 3D autoencoders \cite{groueix1802atlasnet}, \cite{park2019deepsdf}, and image-conditioned 3D reconstruction 
\cite{wang2018pixel2mesh}, \cite{henderson2020leveraging}. 
Hereby, implicit 3D representations have outperformed explicit ones and were extended to learn category-specific 3D morphable models \cite{deng2021deformed}, \cite{zheng2021deep}, and \cite{sun2022topology}.
In contrast, our model does not require synthetic data of several thousands instances per category, but only requires few casually captured object-centric videos. 
This provides novel 3D morphable models for various categories.

\section{A 3D Morphable Model of Neural Features}
\label{sec:method}
In this section we introduce \OURS, a method for self-supervised learning of 3D morphable models (3DMM) for common objects from only object-centric videos.
One of the core challenges that we need to resolve is to find an approach that can deal with highly variable shapes and topologies within the same object category, while at the same time being general enough such that it works for as many categories as possible, hence requiring minimal category-specific heuristics. 
One key design choice is to represent the morphable shape model with a hybrid volumetric-mesh representation that is flexible, but also enables using mesh-based regularizers during training (\cref{sec:method_model}).
Our second key design choice is to represent the object appearance in terms of neural features and not RGB colors (\cref{sec:adapter_img_encoder}), which enables the feature extractor to learn robust 2D-3D correspondences and, hence, facilitates the inverse rendering process.
 We discuss our training strategy and objectives in \cref{sec:method_obj}, and elaborate the inference mechanism in \cref{sec:method_pose}. 
 \cref{fig:method} provides an overview of our method.

\subsection{Morphable Shape Representation}
\label{sec:method_model}

We represent 3D object priors with a category-level template that has neural features attached to its surface, together with instance-specific geometric deformations. 

\textbf{Category-level Template.} One core challenge in our work is to develop a flexible approach for learning the category-level template shape of objects, which generalizes across the highly variable shapes and topologies of common objects.
While purely mesh-based representations can have advantages, they are difficult to optimize from scratch, leading to problems that often require ad-hoc heuristics such as re-meshing \cite{goel2022differentiable,yang2021lasr}. 
To enable a more flexible way of optimizing the category-level 3D shape, we use a hybrid volumetric-mesh representation. 
This combines the advantages of an implicit and explicit 3D representation. 
In particular, the object shape is represented with an implicit model that is given as a signed distance field $\sdff$, which can adapt to complex geometries flexibly. 
It can be converted on the fly into an explicit mesh via the Differentiable Marching Tetrahedra (DMTet) algorithm \cite{shen2021deep}.
Specifically, the signed distance field (SDF) is evaluated at the vertices of a tetrahedral grid and converted into a watertight mesh in a fast and differentiable manner.
We denote the triangle mesh as $\mesh = \{\verts, \edges, \feats \}$ with the vertices $\verts{} = \{\vi \in \mathbb{R}^3\}_{i=1}^{|\verts|}$, the edges $\edges = \{ (\vi,\vj)_{e} \in \mathbb{R}^3 \}_{e=1}^{|\edges|}$, and the features $\feats=\{\feat_{i}\in\mathbb{R}^D\}_{i=1}^{|F|}$.
Once converted to a mesh, the 3D object prior can be rendered into a 2D feature map by applying standard rasterization.
The conversion of the SDF into a mesh enables us to use mesh-based regularizers when learning the deformable model, for example to enforce rigidity.
Moreover, the vertices provide us direct access to corresponding points among the deformed templates, which will enable us to learn features that represent 2D-3D correspondences effectively during training (see \cref{sec:adapter_img_encoder} and \cref{sec:method_obj} for details).

\textbf{Instance-Level Deformation.}
We adopt an affine transformation field \cite{zheng2021deep} to model the deformations of the category-level shape to different object instances.
In contrast to \cite{zheng2021deep} we transform the vertices of the category-level mesh instead of the signed distance field. This is more efficient, as re-extracting the mesh for each instance is not required. 
The function $\affinef(\vv, \instlatent): \mathbb{R}^3 \times \instlatentspace \mapsto \mathbb{R}^3$ maps each vertex $\vv$ from the category-level canonical space together with the image-specific latent code $\instlatent$ to a new position
\begin{equation}
    \begin{aligned}
         \affinef(\vv, \instlatent) = \affinefscale(\vv, \instlatent) \odot \vv + \affineftransl(\vv, \instlatent), \quad \affinefscale, \affineftransl \in \mathbb{R}^3
    \end{aligned},
\end{equation} with the intermediate variables for scale $\affinefscale$ and translation $\affineftransl$. Hereby, $\affinefscale$ and $\affineftransl$ are the outputs of an MLP conditioned on the vertex position $\vv$ and the latent code $\instlatent$. 
We compute the image-specific latent code $\instlatent$ with an image feature encoder $\featencl$ that consists of a DINOv2 backbone and additional convolutional layers (see \cref{fig:method}).
We denote the deformed mesh $\meshdef(\imagesymbol) = \{\vertsdef(\imagesymbol), \edges, \faces, \feats \}$ with the deformed vertices $\vertsdef(\imagesymbol)=\{\affinef(\vi, \featencl(\imagesymbol)) \in \mathbb{R}^3\}_{i=1}^{|\verts|}$ and the image $\imagesymbol$.

\subsection{Appearance as Semantic Feature Field}
\label{sec:adapter_img_encoder}

Prior works in the context of 3DMMs traditionally represent the object appearance using RGB colors, which makes it inherently difficult to apply them for solving vision tasks through inverse rendering \cite{schonborn2017markov,tewari2017mofa,romaszko2017vision}.
A core contribution of our work is to extend 3DMMs to facilitate the inverse rendering process. 
For this purpose, we represent the appearance using neural features instead of RGB colors.
The key advantage of this approach is that the neural features can be invariant to image-specific appearance details and instead focus on encoding semantic correspondences that facilitate the inverse rendering process.
Our approach is inspired from prior works \cite{wu2023magicpony} that learn deformable 3D models through inverse rendering on the feature-level.
These works use DINO \cite{oquab2023dinov2} features that allow estimating semantic correspondences reasonably well \cite{caron2021emerging}.
However, recent studies have shown that DINO features also have limitations, e.g., for differentiating similar object parts, or  \cite{zhang2024telling} or in terms of dense prediction \cite{zhang2024tale}.
Therefore, we choose to learn an adapter on top of the DINO backbone to further improve the image features for better capturing semantic correspondences. 
Specifically, we compute a feature map of semantic features as $S = \Phi_w(\tau(I))  \in \mathbb{R}^{D \times H\times W}$, where $\tau$ is a frozen DINOv2 \cite{oquab2023dinov2} backbone and $\Phi_w$ is the adapter with $w$ being the parameters of several convolutional layers (see \cref{sec:exp_details} for the detailed architecture). 
The adapter is trained jointly with the shape model as described in \cref{sec:method_obj}.

On the 3DMM, we represent the object appearance as a set of feature vectors at every mesh vertex $\feats=\{\feat_{i}\in\mathbb{R}^D\}_{i=1}^{|V|}$. 
The features are generated by a feature field \cite{oechsle2019texture} $\feat_{i} = \featf(\vertex_i): \mathbb{R}^3 \mapsto \mathbb{R}^\featdim$ that maps each vertex $\vertex_i$ to a feature vector $\feat_i$. 
Importantly, as the vertices of the deformable mesh encode corresponding points in different object instances, the adapter can learn more accurate 2D-3D correspondences compared to self-supervised methods that learn purely in 2D, see \cref{fig:method}.

\subsection{Training Objectives}
\label{sec:method_obj}
We train \OURS from a collection of videos of objects.
Our training process jointly optimizes the category-level template shape, the deformation model, the image feature encoder adapter, and the semantic feature field. 
Initially, we obtain a 3D point cloud from every video using Structure-from-Motion \cite{schoenberger2016sfm}.
We need to align the individual 3D reconstructions of object instances into a canonical coordinate frame to obtain a camera pose annotation $\pose$ relative to the object that is coherent across videos.
We achieve this by adopting the method proposed in \cite{sommer2024unsupervised} to get initial camera pose annotations in an unsupervised manner.
In the following, we discuss the geometric objectives, the appearance objective, and finally, the combined objective used for training the model. 

\textbf{Geometric Objectives.}
For optimizing the geometry, we formulate 2D reconstruction objectives using the mask and a 3D reconstruction objective using the 3D point cloud from each instance. Simultaneously, we regularize the category-level template's SDF and introduce mesh-based regularizers for highly rigid instance-level deformations.

First, we penalize the pixel-wise deviation ($u, v$) of our rendered mask $\maskpred(\meshdef, \imagesymbol, \pose, u, v)$ with the pose $\pose$ from the object-centric mask $\mask$, provided by PointRend \cite{kirillov2020pointrend}. For this, we compute the mean squared error
\begin{equation}
    \begin{aligned}
         \lossmask(\meshdef, \imagesymbol, \pose, \mask, u, v) =  \\
         || \maskpred(\meshdef, \imagesymbol, \pose, u, v) - \mask(u, v) ||^2.
    \end{aligned}
\end{equation}
We also maximize the pixel-wise overlap with the mask's distance transform $\maskdt(u, v)$, as follows: 
\begin{equation}
    \begin{aligned}
        &\lossmaskdt(\meshdef, \imagesymbol, \pose, \maskdt, u, v) = \\
        &- \maskpred(\meshdef, \imagesymbol, \pose, u , v) \maskdt(u, v).
    \end{aligned}
\end{equation}
Hereby, the distance transform equals the distance to the silhouette for each pixel inside the mask, all pixels outside the mask are equal to zero. Note that this objective may only increase the object's volume, leading to a category-level template that encapsulates all objects like a visual hull \cite{laurentini1994visual}. 
Without this objective, and given the conflicting objectives of fitting a single category-level template to multiple different instances, the SDF might result in disconnected parts.

To ensure that the deformed mesh reconstructs the shape of the individual object instances, we formulate a 3D chamfer distance based on the canonicalized 3D point cloud $\pointsinst$, obtained from our object-centric video, 
\begin{equation}
\begin{aligned}
&\lossrec(\meshdef, \imagesymbol, \pointsinst) =
     \frac{1}{|\vertsdef| + | \pointsinst|} \\
     &\left( \sum_{\vv_i \in \vertsdef} || \vi - x_{\vnneuc(\vv_i)}||_2 + \sum_{x_i \in \pointsinst} || x_i - \vv_{\vnneuc(x_i)}||_2 \right),
\end{aligned}
\end{equation}
with $\vnneuc: \mathbb{R}^3 \mapsto \mathbb{R}^3$ returning the nearest neighbor of a 3D point to either the vertices $\vertsdef$ or the point cloud $\pointsinst$.

Further, we optimize the category-level template using the Eikonal regulariation as proposed in \cite{gropp2020implicit} to ensure the satisfaction of the Eikonal property of a signed distance function, which states that the norm of the gradient equals to one. This is required as DMTet only provides gradients via the extracted mesh close to the surface, which leaves the points more far away from the surface undefined. Therefore, we sample a set of points $\pointssdf$ uniformly in the canonical space and around the extracted vertices, and average over each point $x \in \pointssdf$
\begin{equation}
    \begin{aligned}
        \losseik(\mesh, x) = (|| \nabla \sdff(x) ||_2 - 1)^2.
    \end{aligned}
\end{equation}
We also penalize large deformations from the category-level template using the mean squared error 
\begin{equation}
    \begin{aligned}
        \lossregdef(\mesh, \meshdef, \imagesymbol)= \frac{1}{|\verts|}\sum_{\vv \in \verts } || \vv - \affinef(\vv, \featencl(\imagesymbol)) ||^2
    \end{aligned}.
\end{equation}
To ensure that deformations of connected vertices are smooth, we adopt the point-pair regularization from \cite{zheng2021deep} to only consider connected vertices
\begin{equation}
    \begin{aligned}
        &\lossregdefsm(\mesh, \meshdef, \imagesymbol) = \frac{1}{|\edges|}  \\ &\sum_{\vi, \vj \in \edges}  \frac{ ||\left[ \vi - \affinef(\vi, \featencl(\imagesymbol)) \right] - \left[ \vj - \affinef(\vj, \featencl(\imagesymbol)) \right] ||_2}{||\vi - \vj||_2}
    \end{aligned}.
\end{equation}

\textbf{Appearance Objective.}
In addition to the morphable shape representation, we also train the parameters of the semantic feature field
$\featf(\vertex_i)$ and the parameters of the adapter module $\Phi_w$ of the image feature encoder.
In order to learn the semantic feature field and the feature encoder simultaneously, we cannot directly use a reconstruction objective, as this optimization collapses to the trivial solution of a constant feature for the whole image and the whole surface. 
We prevent a collapse to the trivial solution by instead optimizing an objective that encourages unique 2D-3D correspondences to emerge, which is inspired from related work on contrastive learning for keypoint detection \cite{bai2023coke}.
In particular, we model for each image feature $s_i$ the 3D correspondence as surface probability $\probpred$, which is computed by the similarity to all vertex features $F$. 
Moreover, we add a background feature $\beta$, which enables our model to differentiate pixelwise between surface and background
\begin{equation}
    \begin{aligned}
            \probpred(\featj | s_i, \feats, \featb) = \frac{\exp \left(\kappa \langle s_i, \featj \rangle \right)}{\sum_{k} \exp \left(\kappa \langle s_i, \featk \rangle \right) + \exp \left(\kappa \langle s_i, \featb \rangle \right)}, 
    \end{aligned}
\end{equation}
where $\kappa$ is the temperature parameter. 
Similarly, the background probability $ \probpred(\featb | s_i, \feats, \featb)$ is defined for each image feature.
In contrast to previous work, we use farthest point sampling as in \cite{qi2017pointnet++} to retrieve a subset of vertices and features. 
To mitigate additional notation, we neglect this sampling in the following. 
As ground truth, we model the surface probability $\prob$ over the set of vertices as in \cite{wang2021nemo,neverova2020continuous}. 
Therefore, the surface probability evaluated at the vertex $\vi$ is defined by the Euclidean distance to all other vertices 
\begin{equation}\label{eq:surface_distribution}
    \begin{aligned}
        \prob(\vj | \vi,\verts) = \frac{\exp \left(- \frac{||\vi - \vj||^2}{2 \sigma^2}\right)}{\sum_{k} \exp \left( \frac{||\vi - \vk||^2}{2 \sigma^2} \right)}
    \end{aligned}.
\end{equation}
Now we can render this surface probability using the camera pose $\probrend(\vj | \meshdef, \pose, u, v)$. For pixels outside the surface we set the background probability $\probrend(\featb | \meshdef, \pose, u, v)$ to 1. Jointly, we estimate these probabilities $\probpred(\featj |  \Phi_w(\imagesymbol, u, v), \feats, \featb)$. 
Thereupon, we optimize the pixel-wise cross-entropy loss
\begin{equation}
    \begin{aligned}
        \lossapp(\Phi_w, \meshdef, \imagesymbol)) = \\
        - \sum_j \log(\probpred(\featj)) \probrend(\vj) - \log(\probpred(\featb)) \probrend(\featb)
    \end{aligned}.
\end{equation}
\textbf{Combined Objective.}
First, we only optimize the category-level template using only the geometric losses 
\begin{equation}
    \begin{aligned}
        \mathcal{L} = \glossrec \lossrec + \glossmask \lossmask + \glossmaskdt \lossmaskdt + \glosseik \losseik,
    \end{aligned}
\end{equation}
with the balancing weights $\lambda_i$.

Second, we optimize the category-level feature field, the adapters, and the instance-level deformations using the weighted sum of geometric and appearance losses
\begin{equation}
    \begin{aligned}
        \mathcal{L} = \glossapp \lossapp + \glossrec \lossrec + \glossmask \lossmask + \glossmaskdt \lossmaskdt \\ 
        + \glosseik \losseik + \glossregdef \lossregdef + \glossregdefsm \lossregdefsm.
    \end{aligned}
\end{equation}

\subsection{Inference via Inverse Rendering}
\label{sec:method_pose}
During inference, our main application is to infer the shape latent $\instlatent$ and the camera pose $\pose$.
The shape latent can be directly predicted using the trained latent encoder $\featencl$.
While the pose could alternatively also be predicted with a feed-forward neural network, other works have demonstrated that inverse rendering can significantly improve the results \cite{wang2022voge}.
Therefore, we optimize the pixelwise surface probabilities during inference applying the reconstruction loss 
\begin{equation}
    \begin{aligned}
        \max_{\pose} \sum_{u, v } \probpred(\featr | \feati, \feats, \featb),
    \end{aligned}
\end{equation}
with the rendered image feature $\featr = \featencr(\meshdef, \pose, u, v)$ and the image feature extracted from the test image $\feati = \Phi_w(\tau(\imagesymbol, u, v))$. Note that we use three degrees of freedom for parametrizing the object rotation. Following \cite{wang2021nemo}, we initialize our pose using uniform sampling.

\section{Experiments}\label{sec:exp}
We conduct extensive experiments and evaluate the effectiveness of \OURS on various downstream tasks for a variety of object categories.
We use the same trained model for all tasks, demonstrating the universal utilization of combining geometry with correspondence-aware appearance in one single model. 
We describe our experimental setup in \cref{sec:exp_details}. Subsequently, we demonstrate that our model improves over prior self-supervised work in 3D object pose estimation (\cref{sec:exp_pose}), in instance segmentation (\cref{sec:exp_seg}), and in semantic correspondence estimation (\cref{sec:exp_sem}).
\subsection{Experimental Details}
\label{sec:exp_details}
\textbf{Training Dataset.}
To train \OURS, we use the object-centric videos provided by the Common Objects in 3D (CO3D) dataset \cite{reizenstein2021common}. For each category, the videos capture a large variety of instances. In total, it provides 19k videos across 50 categories, with each video containing approximately 100 to 200 frames. Each video is supplemented by point clouds and relative camera poses using Structure-from-Motion \cite{schoenberger2016sfm} and object masks using PointRend \cite{kirillov2020pointrend}. 
We choose the same videos as used in \cite{sommer2024unsupervised}, where videos that were captured too narrow to or not focusing on the object were filtered out, and only videos were retained where the full object is visible. 
For each category, we use a maximum of 50 videos. 
Due to the filtering we use fewer videos for the following ten categories: 17 for "remote", 15 for "mouse", 16 for "tv", 7 for "toilet", 41 for "toybus", 28 for "hairdryer", 49 for "couch", and 23 for "cellphone". 
Before training, we center each object's coordinate frame and normalize its scale such that its point cloud ranges from -1 to 1 on the axis with the maximum range. We achieve adopt the method proposed in \cite{sommer2024unsupervised} to obtain canonical camera pose annotations across videos in an unsupervised manner.

\textbf{Evaluation on In-the-Wild Datasets.}
We evaluate our model at various tasks on in-the-wild data.
In particular, we use three common datasets PASCAL3D+ \cite{xiang2014beyond}, ObjectNet3D \cite{xiang2016objectnet3d}, and SPair-71k \cite{min2019spair}. PASCAL3D+ and ObjectNet3D provide pose and segmentation annotations per image, which enables evaluation of 3D object pose estimation as well as instance segmentation. While PASCAL3D+ yields 12 rigid classes of PASCAL VOC 2012, ObjectNet3D covers, beyond that, 100 categories. We use the same 20 common categories across ObjectNet3D and CO3D as used in \cite{sommer2024unsupervised}. 
This even includes the domain gap of a toy bus in CO3D to a real one in PASCAL3D+ and ObjectNet3D. 
In total, we evaluate on 6233 images of PASCAL3D+ using the same subset as in \cite{wang2020robust}, and on 12039 images of ObjectNet3D. We center all objects, as proposed in \cite{zhou2018starmap}. Further, we evaluate our feature adapter on the semantic correspondence task of SPair71k across 6 common categories, which we validate on 3898 image pairs.

\textbf{Implementation Details.} 
As backbone we leverage DINOv2 \cite{oquab2023dinov2} with the ViT-S model \cite{alexey2020image}, for which we freeze all 21M parameters.
To compare with prior work on semantic correspondence estimation \cref{sec:exp_sem}, we additionally trained one ViT-B model. 
From an input resolution of 448x448, the backbone maps to a 32x32 feature map. Our adapter, consisting of three ResNet blocks, upsamples the 32x32 feature map to 64x64. 
For efficient differentiable rendering, we make use of the rasterizer proposed by \cite{Laine2020diffrast}, which achieves differentiability using anti-aliasing. 
For each category-specific morphable model we train, using Adam \cite{kingma2014adam}, for 20 epochs to optimize the template shape, and another 20 to optimize the appearance together with the deformation model. We use early stopping on a hold-out set comprising $10\%$ of the videos. On a single NVIDIA GeForce RTX 2080 the training does not exceed 10 hours. 
\subsection{In-the-Wild 3D Pose Estimation}
\label{sec:exp_pose}
\begin{table}
\centering
\resizebox{\linewidth}{!}{
\begin{tabular}{@{}l|l|rrrrrr|r@{}}
\toprule
 &  Method & \multicolumn{1}{c}{\faIcon{bicycle}} & \multicolumn{1}{c}{\faIcon{bus}}  & \multicolumn{1}{c}{\faIcon{chair}} & \multicolumn{1}{c}{\faIcon{couch}} & \multicolumn{1}{c}{\faIcon{motorcycle}} & \multicolumn{1}{l|}{\faIcon{tv}} & \multicolumn{1}{c}{AVG} \\ \midrule
\multirow{2}{*}{Sup.} & StarMap \cite{zhou2018starmap}       & 83.2                        & 94.4                    & 75.4                      & 79.8                      & 68.8                           & 85.8                    & 82.5                    \\
 & VoGE \cite{wang2022voge}          & 82.6                        & 98.1                    & 90.5                      & 94.9                      & 87.5                           & 83.9                    & 90.9                    \\ \midrule
 \multirow{2}{*}{Unsup.} & ZSP \cite{goodwin2022zero}          & 61.7                        & 21.4                    & 42.6                      & 52.9                      & 43.1                           & 39.0                      & 46.0                      \\
& UOP3D \cite{sommer2024unsupervised}         & 58.4                        & 79.3                    & 51.9                      & 76.6                      & 67.0                             & \textbf{53.1}           & 69.2                    \\
& Ours          & \textbf{67.2}               & \textbf{82.9}           & \textbf{67.8}             & \textbf{80.9}             & \textbf{78.1}                  & 51.1                    & \textbf{75.3}                    \\ \bottomrule
\end{tabular}
}
\caption{\textbf{Evaluation} of 3D object pose estimation on 7 categories of PASCAL3D+, we supplement the missing results in the appendix. The reported metric is the $30^\circ$ accuracy. Note, that ZSP \cite{goodwin2022zero} requires also depth and runs an order of a magnitude slower.}
\label{tab:exp_pose_pascal3d}
\end{table}
\begin{table}
\centering
\resizebox{\linewidth}{!}{
\begin{tabular}{l|rrrrrrr|l}
\toprule
      & \multicolumn{1}{c}{\Backpack} & \multicolumn{1}{c}{\Bench}   & \multicolumn{1}{c}{\faIcon{car}}   & \multicolumn{1}{c}{\faIcon{phone}}  & \multicolumn{1}{c}{\faIcon{chair}}       & \multicolumn{1}{c}{\faIcon{coffee}}    & \multicolumn{1}{l|}{\Hairdryer} &                          \\ \midrule
ZSP \cite{goodwin2022zero}   & 23.1                       & 50.8                      & 60.3                      & 46.4                       & 36.8                         & 33.0                         & \textbf{21.7}                &                          \\
UOP3D \cite{sommer2024unsupervised} & 18.0                         & 62.1                      & 98.1                      & \textbf{54.6}              & 52.2                         & 38.2                       & 14.1                         &                          \\
Ours  & \textbf{25.5}              & \textbf{77.0}               & \textbf{99.0}               & 51.8                       & \textbf{65.9}                & \textbf{44.7}              & 19.4                         &                          \\ \midrule
      & \faIcon{keyboard}                    & \faIcon{laptop}                    & \multicolumn{1}{c}{\Mouse} & \multicolumn{1}{c}{\Remote} & \multicolumn{1}{c}{\faIcon{suitcase}}  & \multicolumn{1}{c}{\Toilet} & \multicolumn{1}{l|}{\faIcon{tv}}      & \multicolumn{1}{r}{AVG}  \\ \midrule
ZSP \cite{goodwin2022zero}   & \textbf{46.8}              & \textbf{60.5}             & 28.8                      & 41.6                       & \textbf{25.8}                & 56.3                       & 37.9                         & \multicolumn{1}{r}{42.2} \\
UOP3D \cite{sommer2024unsupervised} & 26.9                       & 53.3                      & \textbf{44.7}             & \textbf{54.4}              & 15.5                         & 39.6                       & \textbf{53.2}                & \multicolumn{1}{r}{52.4} \\
Ours  & 34.5                       & 54.2                      & 38.5                      & 52.4                       & 15.7                         & \textbf{65.0}                & 52.4                         & \multicolumn{1}{r}{\textbf{56.8}} \\ \bottomrule
\end{tabular}
}
\caption{\textbf{Evaluation} of 3D object pose estimation on 20 categories of ObjectNet3D. We supplement the missing results in the appendix. We report the $30^\circ$ accuracy. Note, that ZSP \cite{goodwin2022zero} requires also depth input and runs an order of a magnitude slower.}
\label{tab:exp_pose_objectnet3d}
\end{table}
We evaluate our 3D object prior at single-image 3D pose estimation on in-the-wild data.
We report the pose estimation results using the $30^\circ$ accuracy in \cref{tab:exp_pose_pascal3d} for PASCAL3D+ and in \cref{tab:exp_pose_objectnet3d} for ObjectNet3D. 
In \cref{tab:exp_pose_pascal3d}, we also report 3D pose estimation results from two supervised methods \cite{zhou2018starmap, wang2022voge} that are trained from the PASCAL3D+ training set.
Notably, all unsupervised methods are trained from CO3D data, as they require videos as input during training, and hence are evaluated in an out-of-domain setting.
Our method improves over prior work on most of the categories except for categories with few shape variance across instances such as cellphone, microwave, and toaster. On average, the $30^\circ$ accuracy improves significantly over prior work from $69.2\%$ to $75.3\%$ by $+6.3$ points on PASCAL3D+ and from $52.4\%$ to $56.8\%$ by $+4.4$ points on ObjectNet3D. We attribute this performance increase to the correspondence learning with a deformable shape, which facilitates the training and, therefore,  generalizes better. This is illustrated in the qualitative examples in \cref{fig:exp_quality}
\subsection{In-the-Wild Instance Segmentation}
\label{sec:exp_seg}
\begin{figure*}
    \centering
    \includegraphics[width=\linewidth,height=6cm]{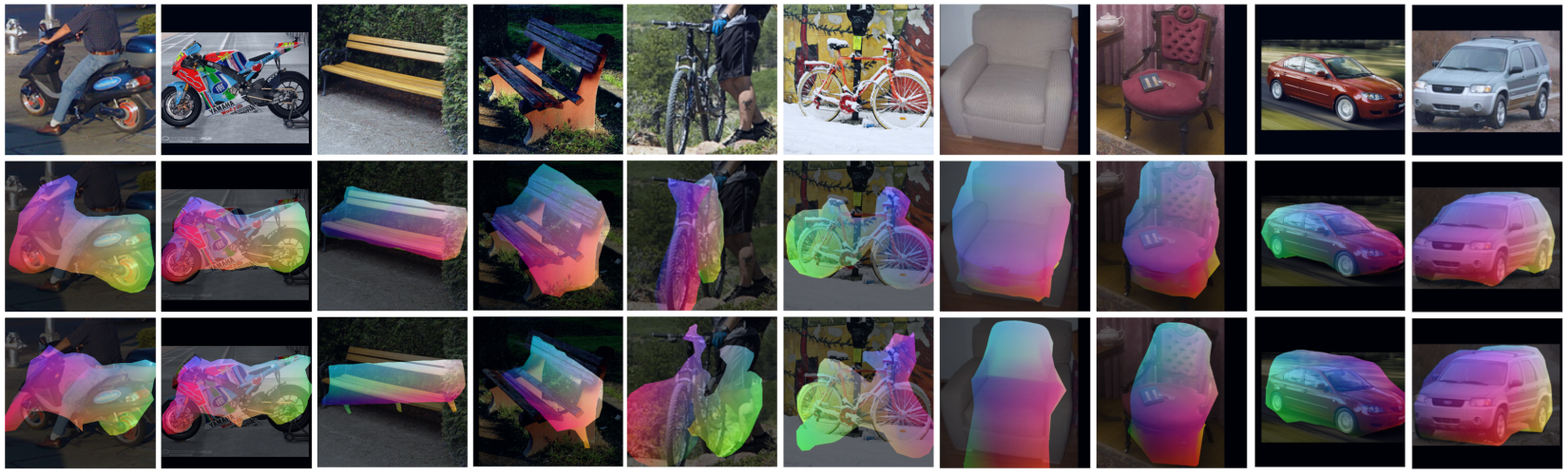}
    \caption{\textbf{Qualitative results} on the ObjectNet3D dataset. In the second row the results of our method are illustrated, in the third the results of UOP3D \cite{sommer2024unsupervised}. Notably, our method fits the object more accurately, resulting in improved 3D pose and segmentation accuracy.}
    \label{fig:exp_quality}
\end{figure*}
\begin{table}[]
\centering
\resizebox{\linewidth}{!}{
\begin{tabular}{@{}l|rrrrrrr|r@{}}
\toprule
 & \multicolumn{1}{c}{\faIcon{bicycle}} & \multicolumn{1}{c}{\faIcon{bus}} & \multicolumn{1}{c}{\faIcon{car}} & \multicolumn{1}{c}{\faIcon{chair}} & \multicolumn{1}{c}{\faIcon{couch}} & \multicolumn{1}{c}{\faIcon{motorcycle}} & \multicolumn{1}{l|}{\faIcon{tv}} & \multicolumn{1}{c}{AVG} \\ \midrule
UOP3D \cite{sommer2024unsupervised}       & 26.0                          & \textbf{66.5}           & 79.1                    & 42.3                      & 71.7                      & 63.6                           & 73.0                      & 60.3                    \\
Ours         & \textbf{29.3}               & 64.7                    & \textbf{82.5}           & \textbf{52.2}             & \textbf{76.9}             & \textbf{65.9}                  & \textbf{77.0}             & \textbf{64.1}           \\ \bottomrule
\end{tabular}
}
\caption{\textbf{Evaluation} of instance segmentation on 7 categories of PASCAL3D+. We report the IoU in percentage with the occlusion-free instance mask provided by the ground truth pose and mesh.}
\label{tab:exp_seg_pascal3d}
\end{table}
\begin{table}[]
\centering
\resizebox{\linewidth}{!}{
\begin{tabular}{l|rrrrrrr|l}
\toprule
      & \multicolumn{1}{c}{\Backpack}  & \multicolumn{1}{c}{\Bench}        & \multicolumn{1}{c}{\faIcon{car}}   & \multicolumn{1}{c}{\faIcon{phone}}  & \multicolumn{1}{c}{\faIcon{chair}}       & \multicolumn{1}{c}{\faIcon{coffee}}    & \multicolumn{1}{l|}{\Hairdryer} &                          \\ \midrule
UOP3D \cite{sommer2024unsupervised} & 68.9                        & 49.9                       & 79.2                      & 67.7                       & 42.9                         & \textbf{69.6}              & 50.9                         &                          \\
Ours  & \textbf{69.4}               & \textbf{54.1}              & \textbf{82.8}             & \textbf{68.1}              & \textbf{53.3}                & 66.8                       & \textbf{53.6}                &                          \\ \midrule
      & \multicolumn{1}{c}{\faIcon{keyboard}} & \multicolumn{1}{c}{\faIcon{laptop}}    & \multicolumn{1}{c}{\Mouse} & \multicolumn{1}{c}{\Remote} & \multicolumn{1}{c}{\faIcon{suitcase}}  & \multicolumn{1}{c}{\Toilet} & \multicolumn{1}{c|}{\faIcon{tv}}      & AVG                      \\ \midrule
UOP3D  \cite{sommer2024unsupervised} & 76.9                        & 62.4                       & \textbf{64.1}             & 61.4                       & 62.5                         & 64.7                       & 73.8                         & \multicolumn{1}{r}{64.0}   \\
Ours  & \textbf{83.1}               & \textbf{70.6}              & 62.7                      & \textbf{71.7}              & \textbf{65.3}                & \textbf{66.7}              & \textbf{78.3}                & \multicolumn{1}{r}{\textbf{67.0}} \\ \bottomrule
\end{tabular}
}
\caption{\textbf{Evaluation} of instance segmentation on 20 categories of ObjectNet3D, we supplement the missing results in the appendix. We report the IoU in percentage with the occlusion-free instance mask given by rendering the annotated mesh in the annotated pose.}
\label{tab:exp_seg_objectnet3d}
\end{table}
We validate the deformation capabilities of \OURS using the instance segmentation masks of PASCAL3D+ in \cref{tab:exp_seg_pascal3d} and ObjectNet3D in \cref{tab:exp_seg_objectnet3d}. 
We compare our model with UOP3D \cite{sommer2024unsupervised},
as it is the current state-of-the-art at self-supervised learning 3D object priors for common objects, and was also trained using CO3D videos.
\OURS outperforms UOP3D on almost all categories and on average increases the IoU by $3.8$ percentage points from $60.3\%$ to $64.1\%$ on PASCAL3D+ and by $3$ percentage points, from $64\%$ to $67\%$ on ObjectNet3D. 
A qualitative comparison of our method against
UOP3D is depicted in \cref{fig:exp_quality}. 
It shows that \OURS achieves much more accurate highly accurate despite a large variability in the object
instances. 
\subsection{Semantic Correspondence}
\label{sec:exp_sem}
\vspace{-.1cm}
\begin{table}[] 
\resizebox{\linewidth}{!}{
\begin{tabular}{@{}l|rrrrrr|r@{}}
\toprule
    & \multicolumn{1}{l}{\faIcon{bicycle}} & \multicolumn{1}{l}{\faIcon{bus}} & \multicolumn{1}{l}{\faIcon{car}} & \multicolumn{1}{l}{\faIcon{chair}} & \multicolumn{1}{l}{\faIcon{motorcycle}} & \multicolumn{1}{l|}{\faIcon{tv}} & \multicolumn{1}{l}{AVG} \\ \midrule
SpherMaps \cite{mariotti2024improving} & 61.2                        & 73.3                    & 67.2                    & 46.3                      & 66.0                          & 42.2                    & 59.4                    \\ \midrule
DINOv2 \cite{oquab2023dinov2}          & \textbf{62.0}                 & 52.3                    & 51.5                    & 36.2                      & \textbf{61.0}                 & 24.2                    & 47.9                    \\
Ours             & 60.3                        & \textbf{60.5}           & \textbf{66.2}           & \textbf{44.2}             & 60.7                        & \textbf{31.9}           & \textbf{54.0}           \\
 \bottomrule
\end{tabular}
}
\caption{\textbf{Evaluation} of the PCK@$0.1$ on 6 categories of SPair-71k \cite{min2019spair}. Note, that SpherMaps \cite{mariotti2024improving} uses coarse viewpoint annotations, while DINOv2 and our approach do not require any. }
\label{tab:exp_sem}
\end{table}
We demonstrate the significant benefit of our semantic feature adapter by evaluating its performance on semantic correspondence estimation in \cref{tab:exp_sem}. 
We evaluate our model compared to the self-supervised DINOv2 model as well as SpherMaps \cite{mariotti2024improving}, which require coarse viewpoint annotations.
We follow the evaluation protocol of SpherMaps  and find correspondences by weighting the image features similarities from the DINOv2 backbone and the feature similarities from our adapter. 
We weight the backbone's features with $0.8$ and ours with $0.2$. On average, we can largely increase the PCK@$0.1$ compared to the self-supervised baseline DINOv2 by $6.1\%$ percentage points from $47.9\%$ to $54\%$. 
For some categories, we can even achieve a competitive performance compared to the supervised baseline SpherMaps.
We believe that one major challenge remains the domain gap between the training and test data.
\subsection{Ablation}
\label{sec:exp_abl}
\vspace{-.1cm}
\begin{table}[]
\centering
\resizebox{\linewidth}{!}{
\begin{tabular}{@{}l|rrr|r@{}}
\toprule
                           & \multicolumn{1}{l}{$30^\circ$} & \multicolumn{1}{l}{$10^\circ$} & \multicolumn{1}{l|}{IoU} & \multicolumn{1}{l}{PCK@0.1} \\ \midrule
w/o Adapt.                 & 25.9                                        & 6.6                                         & 60.5                     & 32.4                        \\
w/o Adapt., w Rec. Loss & 13.9                                        & 2.1                                         & 56.2                     & 32.4                        \\
w/o Dense App. Loss        & 56.7                                        & 12.2                                        & 58.8                     & 45.3                        \\
w/o Inst. Deform.          & \textbf{75.8}                                        & 20.2                                        & 58.6                     & 45.3                        \\
Ours                       & 75.3                                        & \textbf{21.0}                                          & \textbf{64.1}                     & \textbf{47.8}                        \\ \bottomrule
\end{tabular}
}
\caption{\textbf{Ablation} of our model design choices. We evaluate the 3D pose accuracy and instance segmentation on PASCAL3D+ using $30^\circ$, $10^\circ$ accuracy and IoU. Further, the semantic correspondences are evaluated on SPair71-k using PCK@$0.1$. We observe that the adapter is crucial for our method, as well as the dense appearance loss. Further, our instance deformation model yields a minor disadvantage for the coarse $30^\circ$ pose estimation, while improving the fine-grained $10^\circ$ pose estimation, and proves essential for the instance segmentation and the semantic correspondences. Note that in this standard setting we use the ViT-S backbone, in contrast to the results reported in \cref{tab:exp_sem}.}
\label{tab:abl}
\end{table}
In \cref{tab:abl}, we ablate our model design choices regarding the adapter, the appearance loss as well as the instance deformation model. We observe, that instance-level deformation has only a marginal relevance for the pose estimation, but proves essential for segmentation and semantic correspondence estimation. Furthermore, the adapter and the dense appearance loss are significantly improving our method.
\section{Conclusion}
\vspace{-.1cm}
We have introduced \OURS, a new model that can learn 3D morphable models from a collection of casually captured videos completely self-supervised. This model can, at test time, estimate the 3D object shape, 2D-3D correspondences, and the 3D object pose. 
\OURS connects two different subfields of computer vision, self-supervised representation learning and the learning of 3D deformable object priors.
We believe that in the long term, this may unify many vision tasks into a common inference process, while also creating a better 3D spatial understanding in vision models.

\textbf{Acknowledgment.}
This work was funded by the German Research Foundation (DFG) – 468670075, and 499552394 (SFB 1597 - Small Data).

{
    \small
    \bibliographystyle{ieeenat_fullname}
    \bibliography{main}
}


\clearpage
\setcounter{page}{1}
\maketitlesupplementary

\section{Parameters}

\begin{table}[]
\begin{tabular}{l|r}
\hline
Name                        & \multicolumn{1}{l}{Value} \\ \hline
Optimizer                            & Adam                               \\
Learning Rate                        & 1.00E-04                           \\
Batch Size                           & 12                                 \\
Batch Accumulation                   & 2                                  \\
\hline 
Losses                               & \multicolumn{1}{l}{}               \\
$\glossapp$                          & 1.00E-01                           \\
$\glossrec$                          & 1.00E-01                           \\
$\glossmask$                         & 1.00E+00                           \\
$\glossmaskdt$                       & 1.00E+02                           \\
$\glosseik$                          & 1.00E-02                           \\
$\glossregdef$                       & 1.00E-01                           \\
$\glossregdefsm$                     & 1.00E-02                           \\
$\kappa$                             & 14.3                               \\
Vertex Sampling                      & 150                                \\
\hline 
Mesh Model                           & \multicolumn{1}{l}{}               \\
Tetrahedron Grid Size                & 16                                 \\
\hline 
Instance Shape Encoder               & \multicolumn{1}{l}{}               \\
ResNet Blocks                        & 4                                  \\
ResNet Block Types                   & bottleneck                         \\
Out Dimensions                       & {[}256, 256, 256, 256{]}           \\
Strides                              & {[}2, 2, 2, 2{]}                   \\
Pre-Upsampling                       & {[}1, 1, 1{]}                      \\
\hline 
Affine Transformation Field & \multicolumn{1}{l}{}               \\
Layers                               & 5                                  \\
Hidden Dimension                     & 256                                \\
Out Dimension                        & 6                                  \\
\hline 
Feature Adapter                      & \multicolumn{1}{l}{}               \\
ResNet Blocks                        & 4                                  \\
ResNet Block Types                   & bottleneck                         \\
Out Dimensions                       & {[}512, 512, 128{]}                \\
Strides                              & {[}1, 1, 1{]}                      \\
Pre-Upsampling                       & {[}1, 1, 2{]}                      \\
\hline 
SDF Field                            & \multicolumn{1}{l}{}               \\
Layers                               & 5                                  \\
Hidden Dimension                     & 256                                \\
Out Dimension                        & 1                                  \\
\hline 
Feature Field                        & \multicolumn{1}{l}{}               \\
Layers                               & 5                                  \\
Hidden Dimension                     & 256                                \\
Out Dimension                        & 128                                \\ \hline
\end{tabular}
\caption{The following parameters are set for training and testing our model.}
\label{tab:params}
\end{table}

In \cref{tab:params} we list our model's parameters and their values.

We set $\sigma$ as the average nearest neighbor distance for all vertices $\verts$ to the sampled vertices $\verts'$, as follows
\begin{equation}
\begin{aligned}
\sigma = \sqrt{ \frac{1}{|\verts|}\sum_{\vv \in \verts} \min_{\vv' \in \verts'} || \vv - \vv' ||^2}
\end{aligned}.
\end{equation}

\section{Comprehensive Results}

\begin{figure*}
    \centering
    \includegraphics[width=\linewidth]{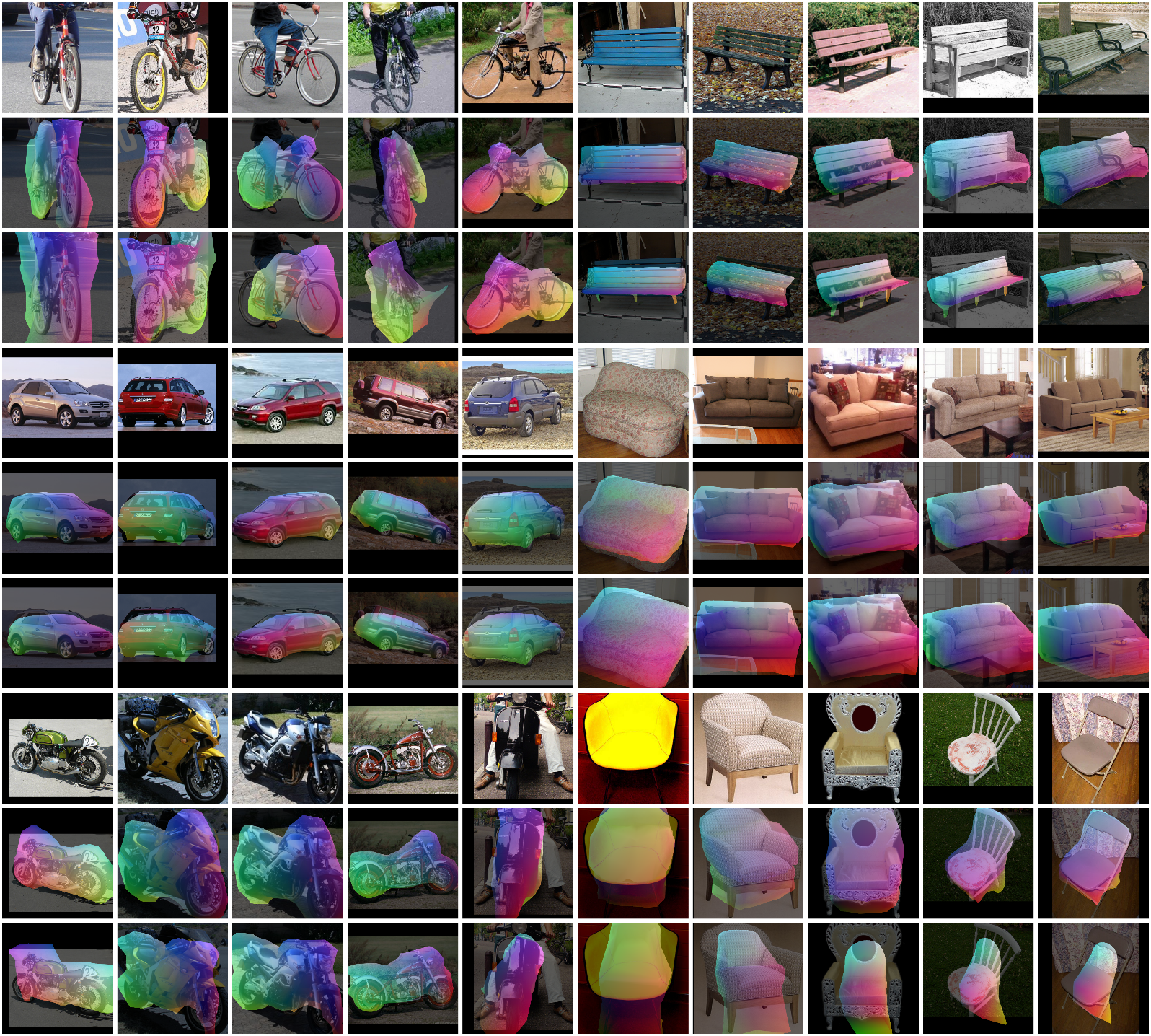}
    \caption{\textbf{Comprehensive qualitative results} on the ObjectNet3D dataset. In the second row the results of our method are illustrated, in the third the results of UOP3D \cite{sommer2024unsupervised}. Notably, our method fits the object more accurately, resulting in improved 3D pose and segmentation accuracy.}
    \label{fig:exp_quality_compr}
\end{figure*}

We report all missing categorical quantitative results in \cref{tab:exp_pose_pascal3d_compr,tab:exp_pose_objectnet3d_compr,tab:exp_seg_objectnet3d_compr}. Additional qualitative results are shown for ObjectNet3D in \cref{fig:exp_quality_compr} and for SPair-71k in \cref{fig:exp_quality_compr_corresp}.

\begin{table*}[]
\centering
\begin{tabular}{@{}l|l|rrrrrrrr@{}}
\toprule
 & Method & \multicolumn{1}{l}{bicycle} & \multicolumn{1}{l}{bus} & \multicolumn{1}{l}{car} & \multicolumn{1}{l}{chair} & \multicolumn{1}{l}{couch} & \multicolumn{1}{l}{motorcycle} & \multicolumn{1}{l}{tv} & \multicolumn{1}{l}{AVG} \\ \midrule
\multirow{2}{*}{Sup.} & StarMap \cite{zhou2018starmap}      & 83.2                        & 94.4                    & 90                      & 75.4                      & 79.8                      & 68.8                           & 85.8                   & 82.5                    \\
& VoGE \cite{wang2022voge}          & 82.6                        & 98.1                    & 99                      & 90.5                      & 94.9                      & 87.5                           & 83.9                   & 90.9                    \\ \midrule 
\multirow{3}{*}{Unsup.} & ZSP \cite{goodwin2022zero}      & 61.7                        & 21.4                    & 61.6                    & 42.6                      & 52.9                      & 43.1                           & 39                     & 46                      \\
& UOP3D \cite{sommer2024unsupervised}         & 58.4                        & 79.3                    & 98.2                    & 51.9                      & 76.6                      & 67                             & \textbf{53.1}          & 69.2                    \\
& Ours          & \textbf{67.2}               & \textbf{82.9}           & \textbf{99.3}           & \textbf{67.8}             & \textbf{80.9}             & \textbf{78.1}                  & 51.1                   & \textbf{75.3}           \\ \bottomrule
\end{tabular}
\caption{\textbf{Comprehensive Evaluation} object 3D pose on PASCAl3D+. We report the $30^\circ$ accuracy for 7 categories.}
\label{tab:exp_pose_pascal3d_compr}
\end{table*}

\begin{table*}[]
\centering
\begin{tabular}{@{}l|rrrrrrrrrr|l@{}}
\toprule
      & \multicolumn{1}{l}{b’pack} & \multicolumn{1}{l}{bench} & \multicolumn{1}{l}{b’cycle} & \multicolumn{1}{l}{bus} & \multicolumn{1}{l}{car}   & \multicolumn{1}{l}{phone}  & \multicolumn{1}{l}{chair}    & \multicolumn{1}{l}{couch}   & \multicolumn{1}{l}{cup}    & \multicolumn{1}{l|}{h’dryer} &                                   \\ \midrule
ZSP \cite{goodwin2022zero}   & 23.1                       & 50.8                      & 58.6                        & 30.5                    & 60.3                      & 46.4                       & 36.8                         & 55.5                        & 33                         & \textbf{21.7}                &                                   \\
UOP3D \cite{sommer2024unsupervised} & 18                         & 62.1                      & 57.8                        & 78.3                    & 98.1                      & \textbf{54.6}              & 52.2                         & 76.6                        & 38.2                       & 14.1                         &                                   \\
Ours  & \textbf{25.5}              & \textbf{77}               & \textbf{65.7}               & \textbf{81.1}           & \textbf{99}               & 51.8                       & \textbf{65.9}                & \textbf{79}                 & \textbf{44.7}              & 19.4                         &                                   \\ \midrule
      & k’board                    & laptop                    & m’wave                      & m’cycle                 & \multicolumn{1}{l}{mouse} & \multicolumn{1}{l}{remote} & \multicolumn{1}{l}{suitcase} & \multicolumn{1}{l}{toaster} & \multicolumn{1}{l}{toilet} & \multicolumn{1}{l|}{tv}      & \multicolumn{1}{r}{AVG}           \\ \midrule
ZSP \cite{goodwin2022zero}  & \textbf{46.8}              & \textbf{60.5}             & 50.5                        & 50.3                    & 28.8                      & 41.6                       & \textbf{25.8}                & 28.8                        & 56.3                       & 37.9                         & \multicolumn{1}{r}{42.2}          \\
UOP3D \cite{sommer2024unsupervised} & 26.9                       & 53.3                      & \textbf{80.3}                        & 69                      & \textbf{44.7}             & \textbf{54.4}              & 15.5                         & \textbf{60.6}               & 39.6                       & \textbf{53.2}                & \multicolumn{1}{r}{52.4}          \\
Ours  & 34.5                       & 54.2                      & 76.1                        & \textbf{81}             & 38.5                      & 52.4                       & 15.7                         & 57.3                        & \textbf{65}                & 52.4                         & \multicolumn{1}{r}{\textbf{56.8}} \\ \bottomrule
\end{tabular}
\caption{\textbf{Comprehensive evaluation} of object 3D pose on ObjectNet3D. We report the $30^\circ$ accuracy for 20 categories.}
\label{tab:exp_pose_objectnet3d_compr}
\end{table*}

\begin{table*}[]
\centering
\begin{tabular}{@{}l|rrrrrrrrrr|l@{}}
\toprule
      & \multicolumn{1}{l}{b’pack}  & \multicolumn{1}{l}{bench}  & \multicolumn{1}{l}{b’cycle} & \multicolumn{1}{l}{bus}     & \multicolumn{1}{l}{car}   & \multicolumn{1}{l}{phone}  & \multicolumn{1}{l}{chair}    & \multicolumn{1}{l}{couch}   & \multicolumn{1}{l}{cup}    & \multicolumn{1}{l|}{h’dryer} &                                 \\ \midrule
UOP3D \cite{sommer2024unsupervised} & 68.9                        & 49.9                       & 27.9                        & \textbf{66.8}               & 79.2                      & 67.7                       & 42.9                         & 73                          & \textbf{69.6}              & 50.9                         &                                 \\
Ours  & \textbf{69.4}               & \textbf{54.1}              & \textbf{31.5}               & 64.9                        & \textbf{82.8}             & \textbf{68.1}              & \textbf{53.3}                & \textbf{77.8}               & 66.8                       & \textbf{53.6}                &                                 \\ \midrule
      & \multicolumn{1}{l}{k’board} & \multicolumn{1}{l}{laptop} & \multicolumn{1}{l}{m’wave}  & \multicolumn{1}{l}{m’cycle} & \multicolumn{1}{l}{mouse} & \multicolumn{1}{l}{remote} & \multicolumn{1}{l}{suitcase} & \multicolumn{1}{l}{toaster} & \multicolumn{1}{l}{toilet} & \multicolumn{1}{l|}{tv}      & AVG                             \\ \midrule
UOP3D \cite{sommer2024unsupervised} & 76.9                        & 62.4                       & \textbf{75.3}               & 64.3                        & \textbf{64.1}             & 61.4                       & 62.5                         & 77.9                        & 64.7                       & 73.8                         & \multicolumn{1}{r}{64}          \\
Ours  & \textbf{83.1}               & \textbf{70.6}              & 75.2                        & \textbf{66.7}               & 62.7                      & \textbf{71.7}              & \textbf{65.3}                & \textbf{78.2}               & \textbf{66.7}              & \textbf{78.3}                & \multicolumn{1}{r}{\textbf{67}} \\ \bottomrule
\end{tabular}
\caption{\textbf{Comprehensive evaluation} of instance segmentation on ObjectNet3D. We report the IoU for 20 categories.}
\label{tab:exp_seg_objectnet3d_compr}
\end{table*}

\begin{figure*}
    \centering
    \includegraphics[width=\linewidth]{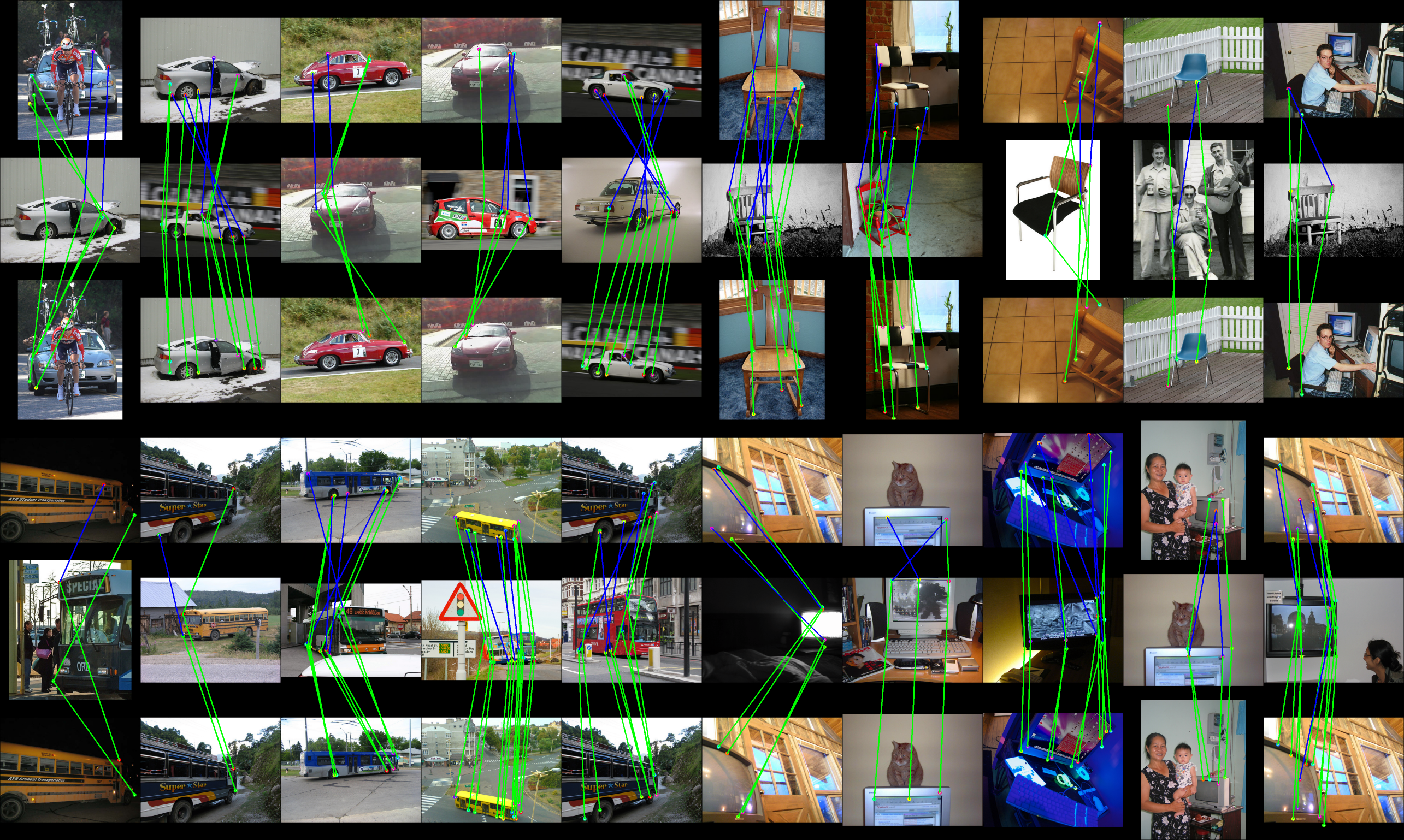}
    \caption{\textbf{Comprehensive qualitative results} on the SPair-71k dataset. In the first row the results of DINOv2 are illustrated, in the third the results of our method. Our method can improve DINOv2 correspondences by resolving ambiguities in parts and symmetries. }
    \label{fig:exp_quality_compr_corresp}
\end{figure*}

\begin{figure*}
    \centering
    \includegraphics[width=\linewidth]{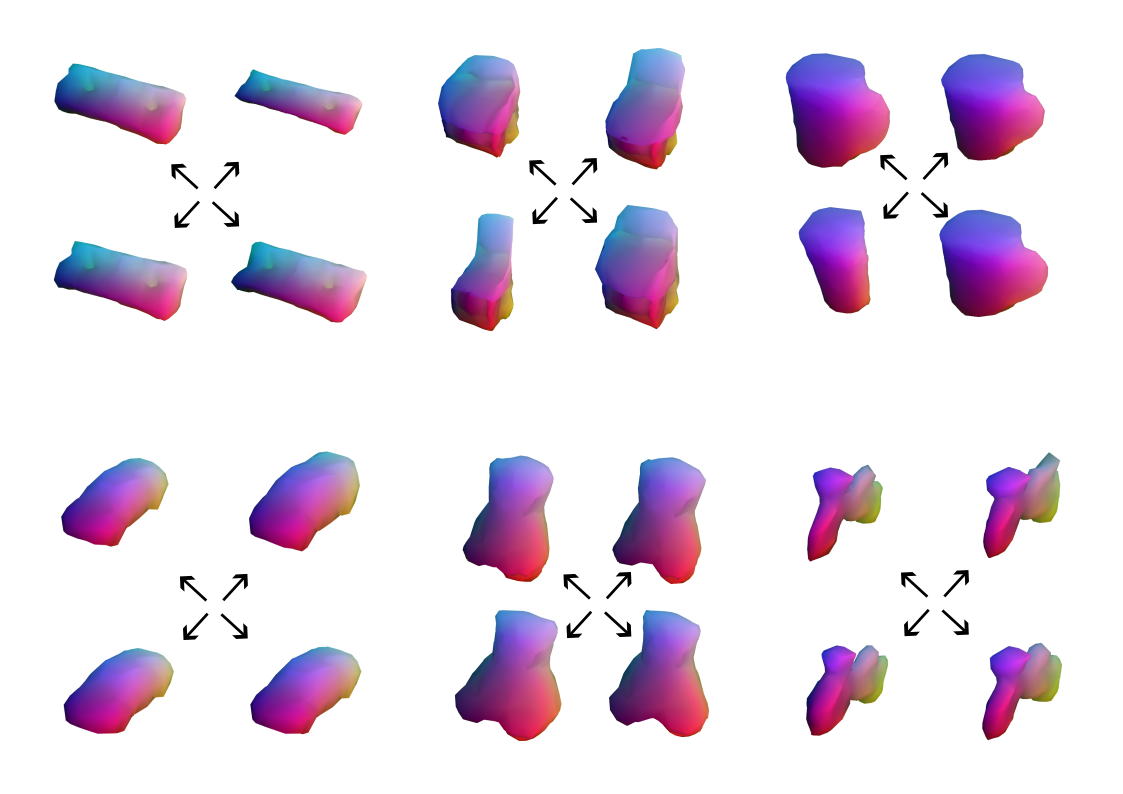}
    \caption{\textbf{3D morphable models with their latent deformations.} }
    \label{fig:exp_quality_compr_corresp}
\end{figure*}


\end{document}